\documentclass[10pt,twocolumn,letterpaper]{article}

\usepackage[pagenumbers]{cvpr} 

\usepackage{cuted} 

\usepackage{currfile} 

\usepackage{caption} 

\usepackage{booktabs}
\usepackage{multirow}
\usepackage{tikz}
\usetikzlibrary{positioning,arrows.meta}

\usepackage{iftex}
\ifXeTeX
  \usepackage{fontspec}
\fi

\definecolor{cvprblue}{rgb}{0.21,0.49,0.74}
\usepackage[pagebackref,breaklinks,colorlinks,allcolors=cvprblue]{hyperref}

\def\paperID{*****} 
\def\confName{CVPR}
\def\confYear{2026}

\title{Ride on Anything: Contact-Coupled Monocular 4D Reconstruction\\of a Rider and Horse in Camera and World Frames}

\author{Zehang Weng\\
{\tt\small zehangweng@gmail.com}
\and
Ci Li\\
{\tt\small cecilialikth@gmail.com}
}

\providecommand{\ProjectLink}{}  
\renewcommand{\ProjectLink}{\par\smallskip\noindent Code and interactive results:\\ \href{https://wengzehang.github.io/rideonanything/}{\texttt{wengzehang.github.io/rideonanything}}}
\begin{document}
\maketitle
\begin{strip}
\centering
\vspace{-1.0em}
\setlength{\tabcolsep}{0pt}
\begin{tabular}{c@{\hspace{4pt}}c@{\hspace{4pt}}c@{\hspace{4pt}}c}
\includegraphics[height=1.02in]{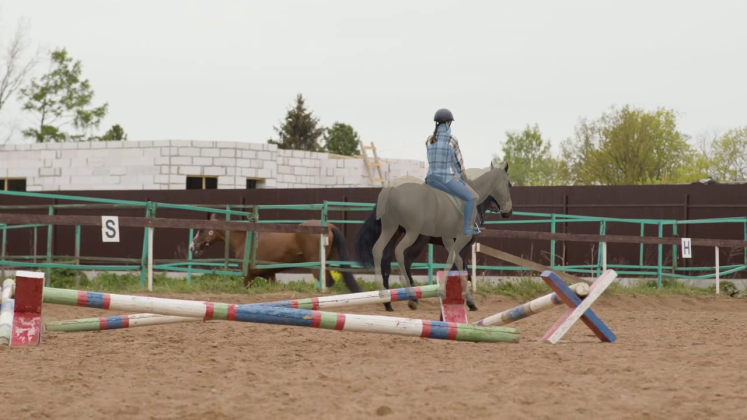} &
\includegraphics[height=1.02in]{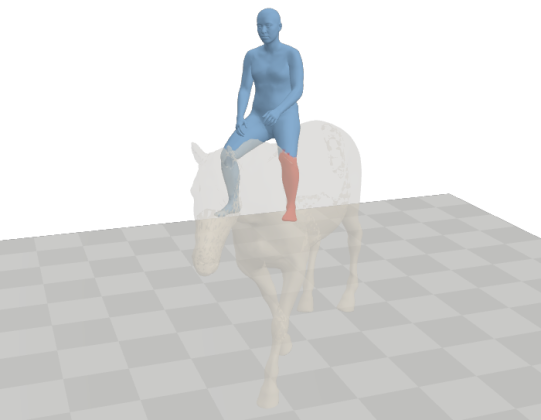} &
\includegraphics[height=1.02in]{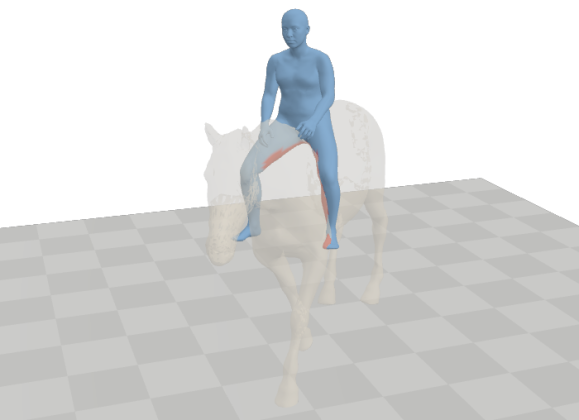} &
\includegraphics[height=1.02in]{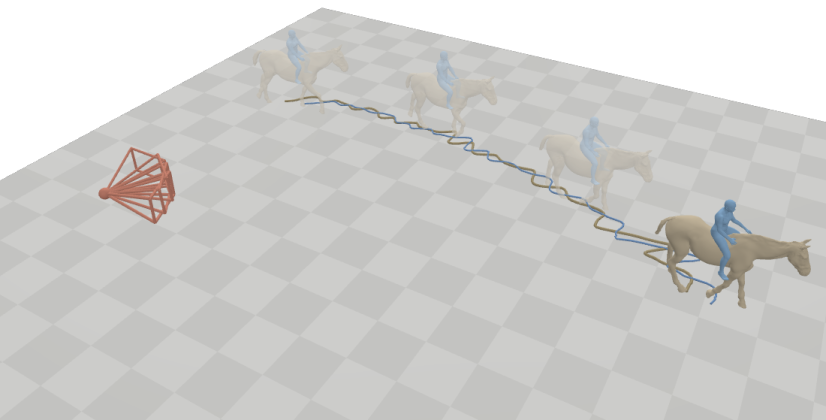}\\[1pt]
\small (a) RoA, input view & \small (b) Baseline, far side & \small (c) RoA, far side & \small (d) RoA, world frame
\end{tabular}
\captionsetup{hypcap=false}
\captionof{figure}{\textbf{Errors that are not visible in the input view} (Walk clip). (a) RoA's reconstruction overlaid on the input frame. (b) The baseline, \ie the raw outputs of the human and horse estimators, seen from the side that the camera does not observe: the rider is not seated and the far leg passes through the horse. Red marks rider surface that lies inside the horse according to the winding-number test; the horse is rendered translucent. (c) RoA in the same view. (d) RoA in a levelled world frame: the horse walks across a static ground while the camera (red) pans to follow it.}
\captionsetup{hypcap=true}
\label{fig:teaser}
\end{strip}

\begin{abstract}
We address the reconstruction of a horse and its rider from a single video. When a human estimator and a horse estimator are applied independently, both meshes can fit their silhouettes well while the rider hovers beside or passes through the horse, slides along its back and jitters relative to it from frame to frame. These errors lie along the camera axis and are not visible in the input view. We present Ride on Anything (RoA), a general, training-free methodology for reconstructing a person riding an animal, which couples a frozen human estimator and a frozen estimator of the mount at test time. We instantiate it for horseback riding with SAM 3D Body and 4DEquine. RoA aligns the outputs of both estimators on a common frame index, projects both subjects through a single pixel-isotropic camera initialized from the horse estimator, and jointly optimizes bounded corrections under interaction terms defined in the horse's body frame, including a collision term based on generalized winding numbers that remains valid on the non-watertight horse mesh. The result is then lifted to a levelled world frame, using the camera rotation estimated from the background and a field of view estimated with MoGe-2. On our riding clips, the raw outputs of the two estimators leave the rider's pelvis 15\,cm away from the horse's back on average. RoA reduces this seat gap to 1\,cm, the drift of the seat along and across the spine by 64\% and 89\%, the 90th-percentile penetration depth by 57\%, and the share of the rider's surface inside the horse from 12.1\% to 7.9\%. The motion is markedly smoother, with 94\% lower acceleration of the rider relative to the horse and 80\% lower horse jerk, while the rider's silhouette agreement decreases from 0.59 to 0.52 IoU. Qualitatively, RoA is substantially better than the baseline: the rider stays seated on the horse and moves together with it. A quantitative evaluation of 3D accuracy awaits calibrated recordings of ridden animals. To probe the limits of the method, we additionally test it on a clip outside its task, a horse led by a person on foot, and analyse the remaining failure modes.
\ProjectLink
\end{abstract}

\section{Introduction}
\label{sec:intro}

Reconstructing a rider and a horse from monocular video involves two articulated bodies that occlude each other and remain in sustained, non-rigid contact. The rider's far leg is hidden behind the barrel and the saddle region is covered by the rider. Recent single-subject estimators perform well for humans~\cite{sam3dbody} and for horses~\cite{4dequine}, but applying them independently to the same video produces meshes that are individually plausible and jointly inconsistent: the rider floats above or sinks into the back, the far leg passes through the horse, and the rider slides along the spine over time. These errors lie mostly along the camera axis. They do not show in an overlay on the input frame and become apparent only when the reconstruction is viewed from another direction (\cref{fig:teaser}).

We present Ride on Anything (RoA), a general, training-free methodology for reconstructing a person riding an animal. It couples a frozen human estimator and a frozen estimator of the mount at test time and reconstructs both subjects in a shared camera frame and in a levelled world frame. The riding-specific parts of RoA, namely the seat on the back, the two sides of the body that the legs straddle, the inside--outside test against the mount and the mount's body frame, are defined on the mount's mesh topology and not on a particular species. Applying RoA to another ridden animal therefore requires a mesh model with fixed topology, the corresponding canonical regions and a pose estimator, but no retraining. In this paper we instantiate and evaluate it for horseback riding, the most common case, with SAM 3D Body~\cite{sam3dbody} and 4DEquine~\cite{4dequine}. The constraints that characterize riding, such as the pelvis resting on the spine, the legs lying on either side of the barrel and no part of the rider being inside the horse, are naturally expressed in the horse's own coordinate frame. When they are expressed on world axes instead and the horse turns, the rider drifts along the spine; on our walking clip it slides 11\,cm backwards. While developing the method we also found that several implementation details have a large effect on the result: a camera scale shared by both image axes, semantic regions selected by slicing along a world axis, the sign convention of the depth axis, and pairing the outputs of the two estimators by list position instead of by video frame. None of these errors is visible in the input view, and we describe them so that they can be avoided.

RoA (i) aligns the two estimators on a common frame index; (ii) projects both subjects through one pixel-isotropic weak-perspective camera initialized from the horse estimator's camera; (iii) evaluates all interaction terms in a per-frame horse body frame, using semantic regions defined once on the canonical horse model; (iv) adds a collision term based on the generalized winding number, which detects rider surface inside the non-watertight horse mesh; and (v) lifts the result to a world frame by estimating the camera rotation from the background and the field of view with MoGe-2~\cite{moge2}.

Because riding videos have no 3D ground truth, we evaluate the reconstructions with metrics that target errors along the camera axis: the share of rider surface inside the horse and its depth, measured with the same winding-number test; the drift of the seat in the horse frame; and world-frame consistency (ground tilt, heading jitter and the speed of stance hooves), in addition to silhouette agreement. We compare RoA with the raw outputs of the two estimators. On the riding clips, these outputs leave the rider 14.9\,cm away from the horse's back on average; RoA reduces this seat gap to 1.1\,cm, the forward and lateral seat drift by 64\% and 89\%, the 90th-percentile penetration depth by 57\%, the share of rider surface inside the horse from 12.1\% to 7.9\%, and the acceleration of the rider relative to the horse by 94\%. The rider's silhouette IoU decreases from 0.59 to 0.52. Qualitatively the improvement is substantial: seen from the side that the camera does not observe, the baseline rider stands beside, hovers above or passes through the horse, whereas RoA keeps the rider seated with the legs on both sides of the barrel and moves the rider together with the horse. Calibrated recordings of ridden animals, which would allow a quantitative evaluation of 3D accuracy, are not yet available. Leading a horse on foot is outside the task that RoA addresses. We nevertheless include one such clip on purpose, to probe the limits of the method: since nobody is mounted, the riding assumption does not hold and RoA, as expected, is worse than the baseline.

Our contributions are:
\begin{itemize}
\item a general, training-free methodology for reconstructing a rider and a ridden animal, which couples off-the-shelf estimators through a shared, correctly scaled camera and interaction terms defined in the mount's body frame, instantiated here for horses;
\item a collision objective and a matching metric based on generalized winding numbers, which handle the non-watertight horse surface;
\item a world-frame lifting for panning cameras with self-checks that indicate when its assumptions are violated;
\item an evaluation on publicly available riding videos with ablations, sensitivity analyses and an analysis of the remaining failure modes. Code and visualization tools will be released.
\end{itemize}

\section{Related Work}
\label{sec:related}

\paragraph{Human mesh recovery.}
Parametric body models such as SMPL~\cite{smpl} and SMPL-X~\cite{smplx} reduce monocular human reconstruction to the estimation of a few hundred parameters, first by fitting to 2D evidence~\cite{smplify} and later by direct regression~\cite{hmr}. SAM 3D Body~\cite{sam3dbody} is a recent promptable full-body regressor built on the MHR body model, and we use it without modification. These methods reconstruct each person in isolation, so the global depth of a rider and the pose of a leg hidden behind the horse are poorly constrained.

\paragraph{Animal and horse reconstruction.}
SMAL~\cite{smal} introduced a shared articulated model for quadrupeds, and hSMAL~\cite{hsmal} specialized it to horses; BARC~\cite{barc} regresses dog shape from images. VAREN~\cite{varen} is an equine body model learned from horse scans, and Dessie~\cite{dessie} estimates articulated horse shape and pose from images with disentangled representations learned from synthetic data. 4DEquine~\cite{4dequine} reconstructs equine motion and appearance from monocular video by predicting VAREN parameters with a temporal model; we use it as our horse estimator. Like human regressors, these methods do not model the rider, who occludes the back and the barrel.

\paragraph{Human--object and multi-body interaction.}
Interacting bodies are commonly reconstructed from a single view by optimizing independently estimated meshes under interaction constraints. PHOSA~\cite{phosa} places humans and objects using depth-ordering, interpenetration and contact terms; PROX~\cite{prox} resolves body pose with the signed distance field of a scanned scene; Jiang \etal~\cite{jiang2020coherent} enforce depth ordering and penalize interpenetration between multiple people. BEHAVE~\cite{behave} and Hi4D~\cite{hi4d} capture human--object and human--human contact with multi-view systems. Riding differs from these settings in several respects: the contact is persistent but non-rigid, both bodies are articulated with very different topologies, the occluded leg lies inside the silhouette of the occluder, and the horse mesh is not watertight, so a signed distance field is not directly available. We use generalized winding numbers~\cite{jacobson2013winding}, an inside--outside test that is robust to such meshes, both in the objective and in the evaluation, and, following~\cite{phosa,jiang2020coherent}, report physical plausibility in addition to image agreement.

\paragraph{World-grounded motion from moving cameras.}
GLAMR~\cite{glamr}, SLAHMR~\cite{slahmr}, WHAM~\cite{wham} and TRAM~\cite{tram} recover human trajectories in a world frame from videos taken with moving cameras, using SLAM, learned motion priors or both. Our world lifting is simpler. Riding videos are typically filmed by a stationary operator who pans the camera, so we estimate only the camera rotation from background homographies~\cite{hartley2004multiple} and obtain the field of view from a monocular geometry model~\cite{moge,moge2}. Translating cameras are left to future work.

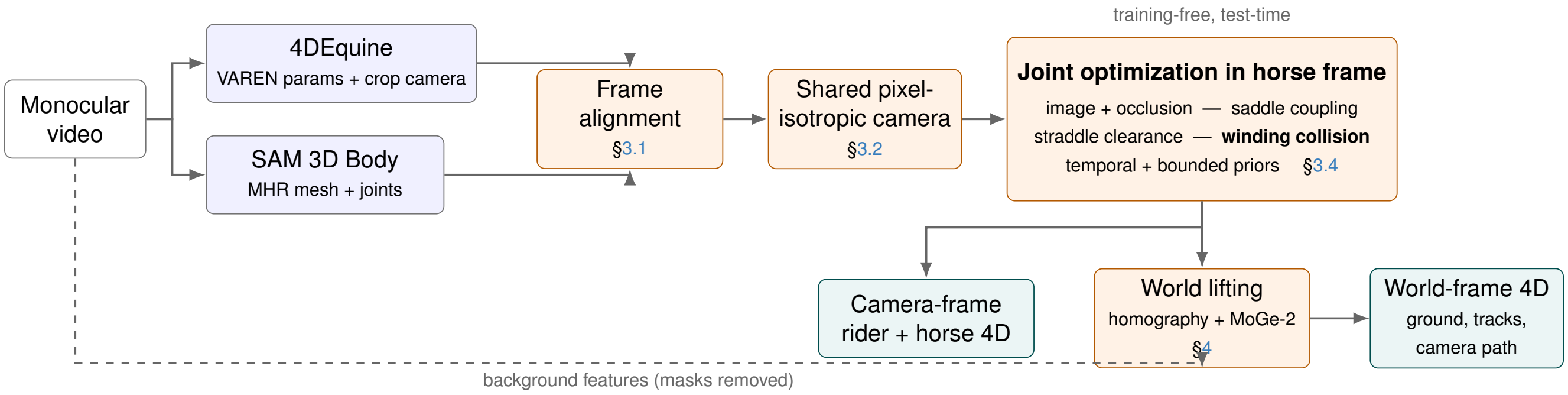
\begin{figure*}[t]
\centering
\resizebox{\linewidth}{!}{%
\begin{tikzpicture}[
  font=\sffamily\small,
  box/.style={draw=black!55, rounded corners=3pt, align=center, inner sep=4pt, minimum height=1.05cm, fill=white},
  model/.style={box, fill=blue!6},
  roe/.style={box, fill=orange!10, draw=orange!70!black},
  outbox/.style={box, fill=teal!8, draw=teal!60!black},
  arrow/.style={-{Latex[length=2.2mm]}, thick, black!60},
  note/.style={font=\sffamily\scriptsize, text=black!60, align=center}]
\node[box, minimum width=1.9cm] (video) {Monocular\\video};
\node[model, right=0.8cm of video, yshift=0.75cm, minimum width=3.2cm] (horse) {4DEquine\\\scriptsize VAREN params + crop camera};
\node[model, right=0.8cm of video, yshift=-0.75cm, minimum width=3.2cm] (human) {SAM 3D Body\\\scriptsize MHR mesh + joints};
\node[roe, right=0.8cm of horse, yshift=-0.75cm, minimum width=2.5cm] (align) {Frame\\alignment\\\scriptsize \S\ref{sec:align}};
\node[roe, right=0.6cm of align, minimum width=2.6cm] (camera) {Shared pixel-\\isotropic camera\\\scriptsize \S\ref{sec:camera}};
\node[roe, right=0.6cm of camera, minimum width=4.4cm, minimum height=2.2cm] (opt) {\textbf{Joint optimization in horse frame}\\[2pt]
  \scriptsize image + occlusion \;|\; saddle coupling\\ \scriptsize straddle clearance \;|\; \textbf{winding collision}\\ \scriptsize temporal + bounded priors \quad \S\ref{sec:objective}};
\node[roe, below=0.9cm of opt, minimum width=2.9cm] (world) {World lifting\\\scriptsize homography + MoGe-2\\\scriptsize \S\ref{sec:world}};
\node[outbox, left=0.8cm of world, minimum width=2.9cm] (camout) {Camera-frame\\rider + horse 4D};
\node[outbox, right=0.8cm of world, minimum width=2.6cm] (worldout) {World-frame 4D\\\scriptsize ground, tracks,\\\scriptsize camera path};
\draw[arrow] (video.east) -- ++(0.35,0) |- (horse.west);
\draw[arrow] (video.east) -- ++(0.35,0) |- (human.west);
\draw[arrow] (horse.east) -| (align.north);
\draw[arrow] (human.east) -| (align.south);
\draw[arrow] (align) -- (camera);
\draw[arrow] (camera) -- (opt);
\draw[arrow] (opt.south) -- ++(0,-0.35) -| (camout.north);
\draw[arrow] (opt.south) -- (world.north);
\draw[arrow] (world) -- (worldout);
\draw[arrow, dashed] (video.south) -- ++(0,-2.75) -| node[note, pos=0.25, below]{background features (masks removed)} (world.south);
\node[note, above=1pt of opt.north] {training-free, test-time};
\end{tikzpicture}}
\caption{\textbf{Overview.} Two frozen estimators reconstruct the horse and the rider independently. RoA aligns their outputs on a common frame index, projects both subjects through one shared camera, and refines them jointly with interaction terms defined in the horse's body frame, including a collision term based on the winding number. The camera-frame result is then lifted to a levelled world frame, using the camera rotation estimated from the background and the field of view estimated with MoGe-2.}
\label{fig:overview}
\end{figure*}

\section{Method}
\label{sec:method}

RoA takes a monocular RGB video and two frozen per-subject estimators, and returns time-aligned rider and horse meshes in the camera frame and in a levelled world frame (\cref{fig:overview}). No training is involved; all quantities are optimized at test time from the estimators' outputs and per-frame instance masks.

\paragraph{Generality.}
The method is formulated for a rider and an arbitrary ridden animal. It requires three mount-specific components: a mesh model with fixed topology and a pose estimator for it; canonical semantic regions for the seat, the two sides of the body and the head and tail, selected once on the rest pose; and the resulting body frame. Everything else, including the alignment, the camera, the objective and the world lifting, is independent of the mount. We describe the method for horses, with VAREN~\cite{varen} as the mesh model and 4DEquine~\cite{4dequine} as the estimator.

\paragraph{Inputs and conventions.}
The horse estimator, 4DEquine~\cite{4dequine}, tracks the horse and predicts, for every tracked frame, the VAREN~\cite{varen} global orientation, articulated pose and shape $\beta^m_t$, together with a weak-perspective camera relative to its square crop. The rider estimator, SAM 3D Body~\cite{sam3dbody}, predicts an MHR mesh $V^h_t$ and 3D keypoints $J^h_t$ for every frame in which a person is detected. Instance masks $M^h_t$ and $M^m_t$ for rider and horse are obtained with a YOLO11 segmentation model~\cite{yolo11}. All geometry is expressed in a right-handed, Y-up scene frame whose $+Z$ axis points towards the camera; this is VAREN's OpenCV output rotated by $180^\circ$ about $X$, and SAM 3D Body already uses this convention. The sign matters because the occlusion term and all far-side metrics depend on which surface is in front.

\subsection{Temporal alignment of the estimators}
\label{sec:align}
The two estimators do not return the same set of frames. The horse tracker discards frames in which the horse is lost and returns the remaining frames concatenated, and the rider detector may miss frames at the clip boundaries. Pairing the $t$-th output of each estimator, or resampling one sequence in time to the length of the other, then pairs rider and horse poses that can be tens of frames apart. RoA therefore records the source frame index of every horse output, keeps the longest contiguous tracked segment of at least 16 frames, and runs the rider estimator and the mask detector on that segment only. It then intersects the rider and horse frame indices and keeps the longest common contiguous run. Meshes, latent parameters, masks and the stored video are all indexed on this timeline, and a remaining mismatch is reported as an error instead of being resampled.

\subsection{A shared pixel-isotropic camera}
\label{sec:camera}
Both subjects are projected through one weak-perspective camera $C_t=(s_t, \mathbf{t}_t)$ per frame. With image aspect ratio $a=W/H$, a scene point $(x,y,z)$ maps to normalized device coordinates
\begin{equation}
u = s_t\,x/a + t^x_t,\qquad v = s_t\,y + t^y_t,
\label{eq:camera}
\end{equation}
so that one scene unit covers the same number of pixels, $k_t=s_tH/2$, in both image directions. Using a single scale for both normalized axes instead stretches each subject horizontally by $W/H$, \ie by $1.78$ for 16:9 video; in our clips the projected silhouettes were then $1.4$ to $1.7$ times too wide.

\paragraph{Initialization from the horse estimator.}
Instead of fitting a camera to a mask bounding box, we convert the horse estimator's crop camera $(\sigma_t,\tau^x_t,\tau^y_t)$, which was predicted for the same VAREN mesh, to the full image. Let $f_c=5000/256\cdot\max(W,H)$ be the crop focal length, $b_t$ the crop size, $\mathbf{c}_t$ the crop centre and $\mathbf{c}_0=(W,H)/2$ the image centre. The full-image depth is $\zeta_t = 2f_c/(b_t\sigma_t)$, and with the offset $\boldsymbol{\delta}_t = 2(\mathbf{c}_t-\mathbf{c}_0)/(b_t\sigma_t)+\boldsymbol{\tau}_t$ we obtain
\begin{equation}
s_t = \frac{2f_c}{\zeta_t H},\qquad
\mathbf{t}_t = s_t\,\mathrm{diag}\big(\tfrac{1}{a},\,-1\big)\,\boldsymbol{\delta}_t,
\end{equation}
which is the weak-perspective limit of 4DEquine's full-image camera; the perspective effect across the horse is below 1\%. The rider is then placed under this camera. A single clip-level scale factor maps the median image height of the rider to metric height, the rider is translated in $X$ and $Y$ to the centre of its mask in each frame, and its depth is set by moving it along the camera axis until the pelvis lies on the centre line of the horse's back in the horse frame (\cref{sec:frame}). In frames where the horse faces the camera this does not constrain the depth, and the clip median is used instead; the per-frame depths are median-filtered.

During optimization the camera remains a variable. It is anchored to its initialization and regularized by its second temporal difference, so that it cannot act as a per-frame zoom that compensates for interaction errors.

\subsection{Interaction in the horse frame}
\label{sec:frame}

\paragraph{Pose-independent semantic regions.}
The interaction terms use regions such as the back, the barrel and the head. Selecting these regions by slicing a posed frame along a world axis works only for a horse seen from the side; for a horse turned by $67^\circ$, such a slice overlapped the true torso with an IoU of only $0.25$. We therefore select the regions once on the canonical VAREN mesh, with identity rotations and the clip-median shape, where the axes are known ($+X$ forward, $+Z$ up). The torso spans 16--58\% of the body length; the back is the top 18\% of the torso and the barrel its upper 62\%; the head and the tail are the front 14\% and the rear 10\% of the body. Since the VAREN topology is fixed, these vertex indices are valid in every frame.

\paragraph{Horse body frame.}
For every frame we construct a body frame $(\mathbf{o}_t,R_t)$ from the torso. The origin $\mathbf{o}_t$ is the torso centroid, the forward axis is the principal direction of the torso oriented towards the head and averaged over a 7-frame window, the up axis is the scene's up axis ($+Y$) orthogonalized against it, and the lateral axis completes a right-handed frame. All interaction variables and terms below are evaluated in the local coordinates $\tilde{\mathbf{x}}=R_t^\top(\mathbf{x}-\mathbf{o}_t)$, and only the image terms are evaluated in the scene frame. Constraints such as ``the legs lie on either side of the barrel'' or ``the pelvis lies on the spine'' are correct on world axes only while the horse moves parallel to the image plane. When the horse turns, they act along a mixture of its lateral and longitudinal directions, which lets the rider slide along the back and pushes the far leg into the barrel. Both effects occur along the camera axis and are not visible in the input view.

\subsection{Variables and objective}
\label{sec:objective}
Rider and horse shapes and the MHR scale parameters are identity attributes, so we fix them to their clip medians. The optimized variables are small per-frame corrections, bounded with $\tanh$ or sigmoid functions: the rider root translation (at most 18\,cm), the horse root translation (4\,cm), a seat offset applied around the pelvis (5.5\,cm), a back offset (4.5\,cm), a lateral barrel scale ($\pm 10\%$), a per-leg abduction about the hip around the horse's forward axis (0.72\,rad), a per-leg lateral offset (16\,cm), and the camera $C_t$. The legs are moved through fixed, smooth weights derived from the hip, knee and ankle joints; individual vertices are not edited. We minimize
\begin{equation}
\begin{aligned}
\mathcal{L} ={} & \underbrace{\lambda_{\rm img}\mathcal{L}_{\rm img} + \lambda_{\rm occ}\mathcal{L}_{\rm occ}}_{\text{image}}
 + \underbrace{\mathcal{L}_{\rm seat} + \lambda_{\rm str}\mathcal{L}_{\rm str} + \lambda_{\rm col}\mathcal{L}_{\rm col}}_{\text{interaction (horse frame)}}\\
 & + \mathcal{L}_{\rm tmp} + \mathcal{L}_{\rm prior}.
\end{aligned}
\end{equation}

\paragraph{Image terms $\mathcal{L}_{\rm img}$ and $\mathcal{L}_{\rm occ}$.}
For the rider, the horse and their union, projected surface samples (\cref{eq:camera}) are compared with the corresponding mask through three quantities: the projected extent against the mask bounding box, an outside-mask distance map sampled at the projected points, and the coverage of the mask contour. All distances are measured isotropically in pixels. The samples are area-weighted because the VAREN head is tessellated much more densely than the barrel. $\mathcal{L}_{\rm occ}$ is a differentiable low-resolution z-buffer. Both subjects are splatted with Gaussian kernels, the front layer is selected with a logistic depth test $\sigma((z_h-z_m)/\tau)$, where a larger $z$ is closer to the camera, and the visible rider and horse occupancies are compared with the masks using binary cross-entropy. Among the image terms, it is the only one that indicates which leg is behind the horse.

\paragraph{Saddle coupling $\mathcal{L}_{\rm seat}$.}
We select one pair of rider-pelvis and horse-back vertices from the initialization and penalize (i) the deviation of the distance between pelvis and back from a target of 2.5\,cm, (ii) the deviation of the offset between the two vertices from a single clip-level offset, split into components normal and tangential to the back surface, (iii) the frame-to-frame change of this offset, and (iv) the second difference of the relative position of the two subjects. Using a fixed offset in the saddle frame, rather than the nearest point in each frame, transfers the sway of the horse to the rider without fixing either root. A centre-line term keeps the pelvis above the spine, and a balance term keeps the rider's centre of mass between the withers and the croup; both are evaluated in the horse frame.

\paragraph{Straddle clearance $\mathcal{L}_{\rm str}$.}
Each leg is assigned to one side of the horse at initialization. The assignment is not updated per frame, since a per-frame assignment would allow a leg to cross the barrel. For samples on the thighs we compute a soft signed distance to the corresponding side of the barrel and penalize, among the worst 25\% of samples, clearances below 6\,mm (including penetration) and above 3\,cm. We also keep the thigh centres 5 to 9.5\,cm outside the skin and the knees at least 12\,cm apart in the lateral direction. After optimization, two short correction steps move each thigh outwards through the leg weights until the 25th percentile of its clearance is at least 2\,mm and its median clearance at least 12\,mm, by at most 1.5\,cm per step.

\paragraph{Winding-number collision $\mathcal{L}_{\rm col}$.}
The straddle term considers only the barrel and depends on local surface normals, so it does not detect a shin that has passed through the chest or a leg that lies entirely inside the horse. We therefore add a global inside--outside test. For 900 area-weighted samples $\mathbf{p}_i$ on the rider's lower body, we compute the generalized winding number~\cite{jacobson2013winding} with respect to the complete VAREN mesh,
\begin{equation}
w(\mathbf{p}) = \frac{1}{4\pi}\sum_{f}\Omega_f(\mathbf{p}),
\end{equation}
where $\Omega_f$ is the signed solid angle subtended by triangle $f$~\cite{vanoosterom1983solid}. The winding number is close to 1 inside the surface and close to 0 outside, and it remains well defined although the VAREN mesh has small holes. Every 10 iterations we store $\mathbf{1}_i=[w(\mathbf{p}_i)>\tfrac12]$ and minimize
\begin{equation}
\mathcal{L}_{\rm col} = \frac{1}{N}\sum_i \mathbf{1}_i\,\big(d(\mathbf{p}_i,\mathcal{S}^m)+\epsilon\big),
\end{equation}
where $d(\cdot,\mathcal{S}^m)$ is the distance to the nearest of 4000 area-weighted samples on the horse surface and $\epsilon=5$\,mm. The gradient moves each inside sample towards the nearest point of the horse surface. The same test is used for the penetration metric (\cref{sec:protocol}), so the objective and the evaluation use one definition of a penetrating sample.

\paragraph{Temporal and prior terms.}
$\mathcal{L}_{\rm tmp}$ penalizes the second differences of all per-frame variables and of the camera. It acts on the corrections introduced by RoA, not on the motion predicted by the estimators, so that, for example, the take-off of a jump is preserved. $\mathcal{L}_{\rm prior}$ keeps the corrections small and includes a coarse penalty on rider points inside an ellipsoid around the horse's body. All weights are the same for every clip (\cref{tab:weights}). We optimize all frames of a clip jointly with Adam~\cite{adam} for 300 iterations.

\section{Lifting to a World Frame}
\label{sec:world}

The image constrains the reconstruction in the camera frame, but riding videos are usually filmed by an operator who pans the camera to keep the horse in view, and in camera coordinates such a horse moves very little. We therefore recover an approximate world frame under the assumption that the camera rotates about a fixed centre, as on a tripod or with a nearly stationary operator, and that its focal length is constant.

\paragraph{Camera rotation from the background.}
We track up to 1200 corners between consecutive frames with pyramidal Lucas--Kanade~\cite{lucaskanade}, using only pixels that are at least 15\,px away from both instance masks, and fit a homography $H_t$ with RANSAC~\cite{ransac}. For a camera that only rotates and has intrinsics $K$, $H_t\simeq K\Delta R_t K^{-1}$~\cite{hartley2004multiple}. We normalize $K^{-1}H_tK$ to unit determinant, project it onto $SO(3)$, and chain the increments, $R_t=\Delta R_t R_{t-1}$, to obtain the rotation of each camera relative to the first one.

\paragraph{Field of view.}
The focal length determines both the recovered pan angle and the metric scale of the trajectory, and it is difficult to observe from small, mostly horizontal pans. Minimizing the rotation residual $\|M M^\top - I\|$ with $M=K^{-1}HK$ over the focal length did not give a well-defined interior minimum on any of our clips. We therefore estimate the vertical field of view $\phi$ with MoGe-2~\cite{moge2} on eight evenly spaced frames, use the median, and set $f=H/(2\tan\tfrac{\phi}{2})$; the spread over the eight frames is stored as an indicator of zoom. All five clips were filmed at long focal lengths, with vertical fields of view between $21^\circ$ and $36^\circ$, so a generic $55^\circ$ prior would have more than doubled the recovered pan on one of them (\cref{fig:fov}).

\paragraph{Placement of the subjects.}
The optimized camera gives $k_t=s_tH/2$ pixels per metre and hence a depth $Z_t=f/k_t$. We back-project the pixel of the horse's torso centre to depth $Z_t$, rotate it into the frame of the first camera with $R_t^\top$, and apply the same rigid transform to all rider and horse vertices of that frame, which leaves the relative placement of rider and horse unchanged. The world position of the torso centre is smoothed with a Savitzky--Golay filter~\cite{savitzky1964smoothing} over 0.6\,s to remove frame-to-frame depth noise; the articulation is not smoothed.

\paragraph{Ground plane.}
We take the lowest 1/60 of the canonical VAREN vertices as hoof samples, collect the lowest 10\% of them in every frame, and fit a plane by SVD. Two degenerate cases are common in riding videos. When the horse moves along a nearly straight line, the plane can rotate freely about that line; and noise can produce planes with an implausible tilt. If the hoof points are nearly collinear ($\sigma_2<0.25\,\sigma_1$) or the fitted plane deviates by more than $10^\circ$ from the up direction of the first camera, we keep the dominant direction in the plane and choose the normal closest to that up direction. The scene is then rotated so that the normal points along $+Y$ and translated so that the ground lies at $Y=0$.

\paragraph{Outputs and self-checks.}
The world track contains the world-frame meshes, the camera centre and the camera axes of every frame (\cref{fig:world}). Since no ground truth is available, it also reports quantities that indicate violations of the assumptions: the tilt of the ground relative to the first camera, the horse speed, the heading jitter of the horse body frame, and the horizontal speed of hooves in contact with the ground (stance slip). Stance slip is zero for a correct reconstruction and increases with depth errors, camera translation and errors in the estimated limb motion.

\section{Experiments}
\label{sec:experiments}

\subsection{Data, baselines and protocol}
\label{sec:protocol}

\paragraph{Clips.}
We use five clips from four publicly available stock videos. All clips have a resolution of $1280\times720$; frame counts and rates are given in \cref{tab:main}. Four of them show riding: a walk during which the horse turns by $33^\circ$ while the camera pans by $65^\circ$ (Walk, 10\,fps); two show-jumping approaches in which the horse comes towards the camera (Jump-A, 13\,fps, and Jump-B, 24\,fps); and a longer, denser sampling of the video of Jump-A (Jump-A long, 50\,fps), which RoA selected automatically with the alignment of \cref{sec:align} and which ends with the horse partly outside the frame. The fifth clip, Lead$^\dagger$, shows a horse led by a person walking beside it, with nobody mounted. Leading is not riding and therefore lies outside the task of RoA; we include the clip deliberately as a stress test of what happens when the riding assumption does not hold, and we do not count it towards the results of the method. Walk and Jump-A were used during development. Jump-B and Lead were not used for any design decision or parameter choice, and Jump-A long was used only to debug the data pipeline. Averages are computed over the four riding clips, and Lead is analysed separately (\cref{sec:failures}). Five clips are not sufficient to establish generalization; the purpose of the study is to measure, for each clip, which components matter and how the method fails.

\paragraph{Baselines and variants.}
The \emph{Baseline} is the raw output of the two estimators: each mesh is placed with the camera translation predicted by its own estimator, SAM 3D Body for the rider and 4DEquine for the horse. Since the two estimators assume different focal lengths, we express both with the focal length of SAM 3D Body and scale the depth of the horse's translation by the ratio of the two focal lengths, which leaves the horse's position and size in the image unchanged. No component of RoA is used. Each ablation removes one component from the full method: the joint optimization as a whole (w/o joint opt., \ie only the shared camera, the alignment and the per-frame rider placement of \cref{sec:align,sec:camera}), the contact coupling ($\mathcal{L}_{\rm seat}$), the image terms ($\mathcal{L}_{\rm img}$, $\mathcal{L}_{\rm occ}$), the horse body frame (interaction terms on world axes, as in our earlier implementation), or the winding-number collision ($\lambda_{\rm col}=0$). All runs use the same estimator outputs, masks and weights (\cref{tab:weights}), and all of them are reproduced by a single script included in the code release.

\paragraph{Metrics.}
\emph{Silhouette IoU} (IoU$_h$, IoU$_m$) compares each mesh, rasterized with the camera of the respective stage, with the detected mask. Because the masks are computed automatically, IoU measures agreement with the detector rather than accuracy. \emph{Inside\,\%} is the share of 2500 area-weighted rider surface samples whose winding number with respect to the complete horse mesh exceeds $\tfrac12$, averaged over frames. Penetration must be read together with the \emph{seat gap}, the closest distance between the rider's pelvis and the horse's back vertices in each frame, averaged over the clip: a rider who is far from the horse cannot penetrate it. The \emph{depth} of an inside sample is its distance to the nearest horse vertex; we take the maximum per frame and report the 90th percentile (P90) and the maximum over the clip. A depth of a few centimetres is expected even for a correct reconstruction, since the thighs press into the saddle and soft tissue, neither of which is modelled; larger values indicate a leg inside the horse. \emph{Drift} is the standard deviation over the clip of the pelvis position relative to the horse's back, measured along the forward and lateral axes of the horse body frame. A seated rider is nearly static in this frame, so drift measures sliding along and across the spine, which is not visible from the side. The world-frame quantities are defined in \cref{sec:world}. Several of these metrics are closely related to terms of the objective: drift and relative acceleration to the saddle coupling, and the inside share to the collision term. They measure how well RoA satisfies the riding constraints, not the accuracy of the reconstruction.

\begin{table*}[t]
\centering\small
\caption{\textbf{Main results.} Baseline: the raw outputs of the two estimators, each placed with its own predicted camera. Gap: seat gap between pelvis and back. Depth, gap and drift in cm; bold marks the better of the two rows. $^\dagger$Lead: a horse led by a person on foot. It lies outside the riding task and is included only as a stress test (\cref{sec:failures}); it is excluded from all averages.}
\label{tab:main}
\begin{tabular}{llccccccc}
\toprule
Clip & Method & IoU$_h\uparrow$ & IoU$_m\uparrow$ & Inside\,\%$\downarrow$ & P90$\downarrow$ & Gap$\downarrow$ & Fwd.\ drift$\downarrow$ & Lat.\ drift$\downarrow$\\
 & & & & & (cm) & (cm) & (cm) & (cm)\\
\midrule
\multirow{2}{*}{\shortstack[l]{Walk\\\scriptsize 165f @ 10\,fps}} & Baseline & \textbf{0.667} & \textbf{0.517} & 12.0 & 28.2 & 10.4 & 8.7 & 17.1\\
 & RoA & 0.565 & 0.497 & \textbf{9.4} & \textbf{7.0} & \textbf{0.8} & \textbf{1.2} & \textbf{2.5}\\
\midrule
\multirow{2}{*}{\shortstack[l]{Jump-A\\\scriptsize 40f @ 13\,fps}} & Baseline & \textbf{0.565} & 0.499 & 9.7 & 24.3 & 16.0 & 13.2 & 5.9\\
 & RoA & 0.540 & \textbf{0.540} & \textbf{6.7} & \textbf{9.9} & \textbf{1.0} & \textbf{4.9} & \textbf{1.3}\\
\midrule
\multirow{2}{*}{\shortstack[l]{Jump-A long\\\scriptsize 180f @ 50\,fps}} & Baseline & \textbf{0.539} & 0.503 & 12.0 & 26.9 & 13.5 & 10.1 & 10.5\\
 & RoA & 0.477 & \textbf{0.505} & \textbf{8.8} & \textbf{16.2} & \textbf{1.0} & \textbf{4.6} & \textbf{2.4}\\
\midrule
\multirow{2}{*}{\shortstack[l]{Jump-B\\\scriptsize 103f @ 24\,fps}} & Baseline & \textbf{0.606} & 0.447 & 14.6 & 30.5 & 19.6 & 14.3 & 36.0\\
 & RoA & 0.506 & \textbf{0.474} & \textbf{6.7} & \textbf{13.7} & \textbf{1.4} & \textbf{6.0} & \textbf{1.6}\\
\midrule
\multirow{2}{*}{\shortstack[l]{Lead$^\dagger$\\\scriptsize 125f @ 24\,fps}} & Baseline & \textbf{0.709} & \textbf{0.469} & \textbf{0.0} & \textbf{0.0} & 134.3 & \textbf{9.8} & 51.1\\
 & RoA & 0.406 & 0.396 & 3.4 & 10.5 & \textbf{98.1} & 10.2 & \textbf{7.5}\\
\bottomrule
\end{tabular}
\end{table*}

\subsection{Results}
\label{sec:results}

\paragraph{Comparison with the baseline.}
The raw outputs of the two estimators do not form a valid riding configuration (\cref{tab:main}). Averaged over the riding clips, the rider's pelvis is 14.9\,cm away from the horse's back, and it drifts by 11.6\,cm along and 17.4\,cm across the spine over a clip, because the two estimators predict their depths independently. RoA places the rider on the horse: the seat gap drops to 1.1\,cm ($-93\%$), the forward drift to 4.2\,cm ($-64\%$) and the lateral drift to 1.9\,cm ($-89\%$). Penetration decreases as well: the share of rider surface inside the horse falls from 12.1 to 7.9\% ($-34\%$), the P90 penetration depth from 27.5 to 11.7\,cm ($-57\%$) and the maximum from 31.0 to 13.4\,cm. In the baseline, some part of the rider, often the far leg, lies more than 2\,cm inside the horse in 86 to 99\% of the frames. The seated RoA rider is in contact with the horse in every frame, and most of its remaining penetration is the inner thighs entering the barrel by a few centimetres (see below); only on Walk is the share of the far side inside the horse higher for RoA than for the baseline (16.0 vs.\ 12.8\%).

\begin{figure*}[t]
\centering
\includegraphics[width=\linewidth]{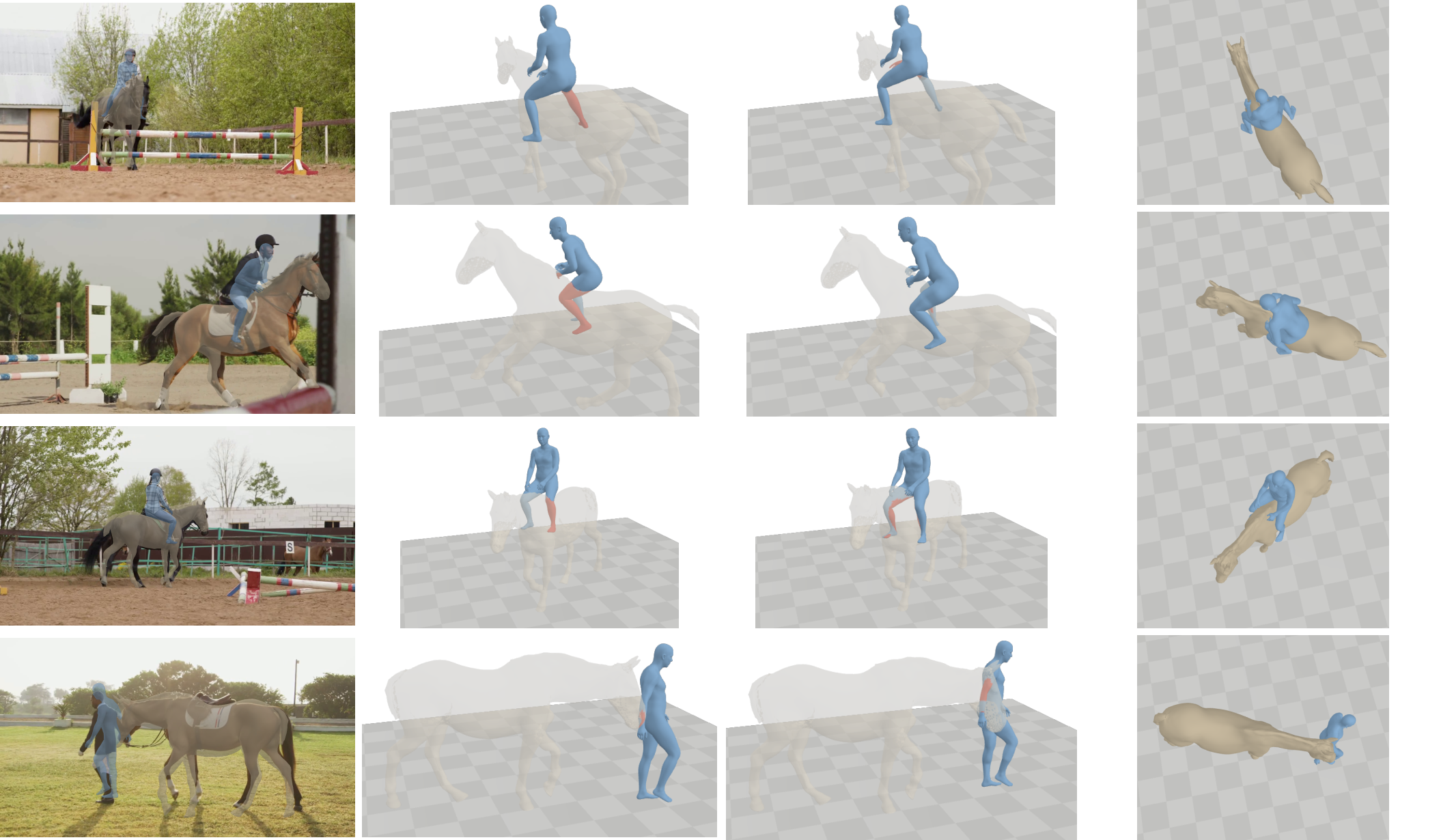}\\[1pt]
{\small\makebox[0.25\linewidth]{(a) RoA, input view}\makebox[0.25\linewidth]{(b) Baseline, far side}\makebox[0.25\linewidth]{(c) RoA, far side}\makebox[0.25\linewidth]{(d) RoA, top view}}
\caption{\textbf{Qualitative results.} Column (a): RoA overlaid on the input frame; (b): the baseline seen from the side opposite the camera; (c): RoA in the same view; (d): RoA from above. Red marks rider surface inside the horse (winding number $>\frac12$), and the horse is rendered translucent. Rows, top to bottom: Jump-A, Jump-B, Walk, and Lead, a stress test outside the riding task in which nobody is mounted and RoA, applying its riding terms, misplaces the person relative to the horse.}
\label{fig:qualitative}
\end{figure*}

\paragraph{Qualitative comparison.}
The qualitative difference is larger than the scalar metrics suggest. Seen from any direction other than the camera's, the baseline rider stands next to, hovers above or passes through the horse and jumps relative to it from frame to frame (\cref{fig:teaser,fig:qualitative}), whereas RoA keeps the rider seated with the legs on the two sides of the barrel and moves the rider together with the horse. We consider this the most important result. It rests on visual inspection and on the gap, drift and smoothness metrics; a quantitative measure of 3D accuracy requires calibrated recordings of ridden horses, which are not yet available (\cref{sec:limitations}).

\paragraph{Silhouette agreement.}
The baseline has the higher rider IoU on every riding clip (0.594 vs.\ 0.522 on average), while the horse IoU is similar (0.492 vs.\ 0.504). In the baseline each mesh is fitted to the image separately with its own perspective camera, with no requirement that the two be consistent in 3D. Most of the decrease comes from projecting both subjects through one shared weak-perspective camera and placing the rider under it (0.594 to 0.546 without joint optimization, \cref{tab:ablation}); the joint optimization, which moves the rider by a few centimetres to resolve contact and collisions, lowers it further to 0.522.

\paragraph{Temporal smoothness.}
RoA also makes the motion considerably smoother (\cref{tab:smoothness}). The RMS second difference of the pelvis position relative to the horse's back, \ie the jitter of the rider relative to the horse and the most visible artefact during playback, decreases by 90 to 96\% on every riding clip (94\% on average). The horse's jerk, measured as the mean magnitude of the third difference of its centroid, decreases by 73 to 83\% (80\% on average), because the per-frame depth of the baseline comes from the horse estimator's crop camera, whereas RoA's shared camera is temporally regularized. The rider's jerk decreases on three clips, by 27\% on average, but increases on Jump-B, where the rider follows the horse's body, whose heading jitters by $8^\circ$ per frame on this clip (\cref{tab:world}). The corrections that RoA applies to the horse itself are small: its vertices move by 2.3\,cm on average and 99\% of them by at most 5.0\,cm on every clip, and its centroid moves by at most 4.9\,cm, \ie 1 to 2\% of the horse's body length of about 2.5\,m.

\begin{table}[t]
\centering\scriptsize\setlength{\tabcolsep}{3.2pt}
\caption{\textbf{Temporal smoothness} on the riding clips, Baseline (Base.) vs.\ RoA. Rider and horse jerk: mean magnitude of the third difference of the centroid (m/s$^3$); it grows with the frame rate, so values are comparable only within a clip. Rel.\ accel.: RMS second difference of the pelvis offset relative to the horse's back, in the horse frame ($10^{-3}$\,m per frame$^2$). Mean change: mean of the per-clip relative changes. Bold: lower value per clip.}
\label{tab:smoothness}
\begin{tabular}{lcccccc}
\toprule
Clip & \multicolumn{2}{c}{Rider jerk$\downarrow$} & \multicolumn{2}{c}{Rel.\ accel.$\downarrow$} & \multicolumn{2}{c}{Horse jerk$\downarrow$}\\
 & Base. & RoA & Base. & RoA & Base. & RoA\\
\midrule
Walk & 295 & \textbf{64} & 361.0 & \textbf{13.8} & 344 & \textbf{59}\\
Jump-A & 573 & \textbf{203} & 345.6 & \textbf{22.9} & 944 & \textbf{167}\\
Jump-A long & 11837 & \textbf{8828} & 180.2 & \textbf{17.7} & 24266 & \textbf{6603}\\
Jump-B & \textbf{470} & 759 & 361.0 & \textbf{15.1} & 4360 & \textbf{778}\\
\midrule
Mean change & \multicolumn{2}{c}{$-27$\%} & \multicolumn{2}{c}{$-94$\%} & \multicolumn{2}{c}{$-80$\%}\\
\bottomrule
\end{tabular}
\end{table}

\begin{table}[t]
\centering\scriptsize\setlength{\tabcolsep}{2.6pt}
\caption{\textbf{Baseline and ablations}, mean over the riding clips (per clip: \cref{tab:perclip}). Depth, gap and drift in cm. Bold: best value per column over all rows.}
\label{tab:ablation}
\begin{tabular}{lccccccc}
\toprule
Variant & IoU$_h$ & IoU$_m$ & Inside & P90 & Gap & Fwd. & Lat.\\
 & $\uparrow$ & $\uparrow$ & \%\,$\downarrow$ & cm\,$\downarrow$ & cm\,$\downarrow$ & cm\,$\downarrow$ & cm\,$\downarrow$\\
\midrule
Baseline & \textbf{0.594} & 0.492 & 12.1 & 27.5 & 14.9 & 11.6 & 17.4\\
\midrule
w/o joint opt. & 0.546 & 0.494 & 14.6 & 20.3 & 4.4 & 7.9 & 2.2\\
w/o contact & 0.524 & 0.503 & 8.5 & 11.7 & 1.3 & 5.0 & \textbf{1.8}\\
w/o image & 0.453 & 0.484 & 7.3 & 12.0 & \textbf{1.1} & 4.1 & 1.9\\
world axes & 0.519 & \textbf{0.511} & \textbf{5.0} & \textbf{8.0} & 2.0 & \textbf{4.0} & 2.4\\
w/o collision & 0.527 & 0.506 & 11.3 & 15.4 & 1.2 & 4.1 & 1.9\\
\midrule
Full RoA & 0.522 & 0.504 & 7.9 & 11.7 & \textbf{1.1} & 4.2 & 1.9\\
\bottomrule
\end{tabular}
\end{table}

\begin{figure*}[t]
\centering
\includegraphics[width=\linewidth]{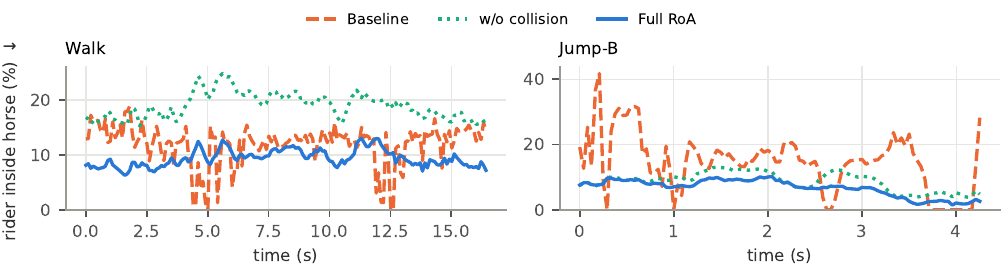}\\[-2pt]
\includegraphics[width=\linewidth]{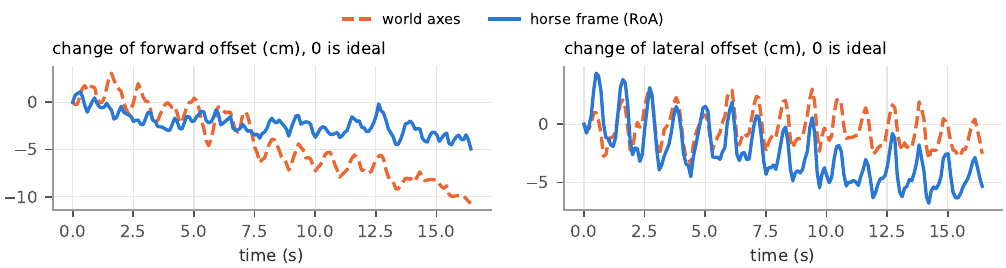}
\caption{\textbf{Per-frame analysis.} Top: share of rider surface inside the horse on Walk and Jump-B (lower is better, given a seated rider). Bottom: change of the pelvis offset relative to the horse's back on Walk, in the horse frame and relative to the first frame (a seated rider should stay near 0); with interaction terms on world axes, the rider slides backwards along the spine as the horse turns.}
\label{fig:curves}
\end{figure*}

\paragraph{Ablations.}
\Cref{tab:ablation} isolates the components of RoA. The shared camera, the alignment and the per-frame rider placement alone (w/o joint opt.) bring the rider close to the horse (seat gap 4.4\,cm), but leave the far leg deep inside it: 14.6\% of the rider's surface is inside, with a P90 depth of 20.3\,cm. The joint optimization reduces this to 7.9\% and 11.7\,cm ($-46\%$ and $-42\%$), the forward drift from 7.9 to 4.2\,cm, and the relative acceleration from 96.8 to $17.4\times10^{-3}$\,m per frame$^2$. Without the winding-number collision, penetration increases to 11.3\% and the P90 depth to 15.4\,cm, while IoU barely changes. Without the contact coupling, the forward drift increases from 4.2 to 5.0\,cm on average, mostly on Jump-B (8.8 vs.\ 6.0\,cm); on the other clips the difference is at most 0.5\,cm. Without the image terms, the rider IoU drops by 0.07, because the interaction terms can then move the rider away from its position in the image. The results for the horse body frame are mixed. On Walk, where the horse turns, world axes let the rider slide 11\,cm backwards along the spine, against 4\,cm in the horse frame (\cref{fig:curves}, bottom). On the jump clips, where the horse faces the camera, world axes give lower forward drift and lower penetration (5.0\% vs.\ 7.9\% on average), since they push the legs along the unobserved camera axis, which can take a leg out of the barrel without placing it beside the horse; far-side ground truth would be needed to decide which is more accurate.

\begin{table}[t]
\centering\scriptsize\setlength{\tabcolsep}{3pt}
\caption{\textbf{Collision weight} on Walk. Time: joint optimization only, one RTX~3090. Bold: best per column.}
\label{tab:sensitivity}
\begin{tabular}{lccccc}
\toprule
$\lambda_{\rm col}$ & IoU$_h\uparrow$ & IoU$_m\uparrow$ & Inside\,\%$\downarrow$ & P90 (cm)$\downarrow$ & Time (s)$\downarrow$\\
\midrule
0 & 0.567 & \textbf{0.513} & 19.0 & 18.9 & \textbf{121}\\
100 & 0.565 & 0.500 & 12.0 & 9.5 & 233\\
400 & 0.565 & 0.497 & 9.4 & 7.0 & 236\\
1500 & 0.569 & 0.495 & 7.5 & 5.5 & 233\\
4000 & \textbf{0.578} & 0.494 & \textbf{6.8} & \textbf{4.6} & 233\\
\bottomrule
\end{tabular}
\end{table}

\paragraph{Collision weight and runtime.}
\Cref{tab:sensitivity} varies $\lambda_{\rm col}$ on Walk. Penetration decreases from 19.0\% without the term to 9.4\% at the default weight of 400 and to 6.8\% at 4000, and the P90 depth from 18.9 to 7.0 and 4.6\,cm, while the IoUs change by at most 0.02. The collision term doubles the optimization time, from 121 to 233\,s for 165 frames, because the winding number is evaluated against all 27.7k triangles of the horse mesh. Including both estimators, processing a clip of a few seconds takes 5 to 7 minutes.

\begin{table}[t]
\centering\scriptsize\setlength{\tabcolsep}{2.4pt}
\caption{\textbf{World frame.} Vertical FOV from MoGe-2, maximum camera rotation relative to the first frame, horse travel and mean speed, ground tilt relative to the first camera, heading jitter (Jit.) of the horse body frame, and median speed of hooves in ground contact (stance slip). $^\ast$: the ground plane was not determined by the hooves and was constrained to contain the main direction of the hoof track (\cref{sec:world}), so its tilt only reflects the slope of that direction. $\downarrow$ marks self-checks, for which lower values indicate fewer violations of the assumptions; the other columns describe the clip and have no preferred direction.}
\label{tab:world}
\begin{tabular}{lccccccc}
\toprule
Clip & FOV & Pan & Trav. & Speed & Tilt$\downarrow$ & Jit.$\downarrow$ & Slip$\downarrow$\\
 & ($^\circ$) & ($^\circ$) & (m) & (m/s) & ($^\circ$) & ($^\circ$/f) & (m/s)\\
\midrule
Walk & 26.4 & 65 & 15.4 & 0.94 & 0.5$^{\ast}$ & 2.4 & 1.17\\
Jump-A & 21.4 & 5 & 6.9 & 2.34 & 0.6$^{\ast}$ & 2.5 & 2.28\\
Jump-A long & 20.8 & 11 & 9.6 & 2.67 & 0.8$^{\ast}$ & 1.6 & 2.70\\
Jump-B & 23.8 & 10 & 7.3 & 1.71 & 11.6$^{\ast}$ & 8.3 & 1.50\\
Lead$^\dagger$ & 35.5 & 5 & 1.9 & 0.37 & 1.3 & 0.5 & 0.35\\
\bottomrule
\end{tabular}
\end{table}

\begin{figure}[t]
\centering
\includegraphics[width=\columnwidth]{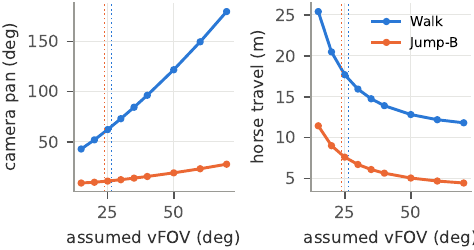}
\caption{\textbf{Effect of the field of view.} Recovered camera pan and horse travel for different assumed vertical fields of view; the dotted lines mark the MoGe-2 estimate. The plot shows the sensitivity of the world trajectory, so neither direction is better. With a generic $55^\circ$ prior, the pan on Walk would be twice as large.}
\label{fig:fov}
\end{figure}

\begin{figure*}[t]
\centering
\includegraphics[width=\linewidth]{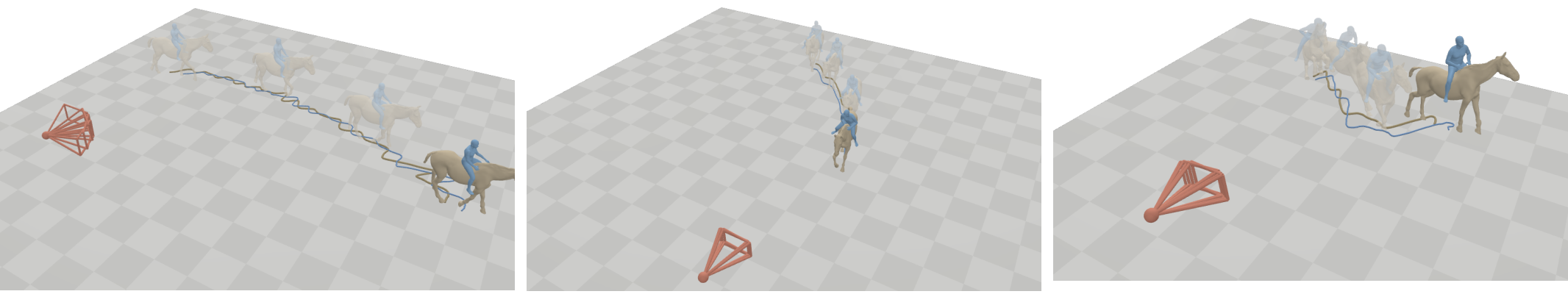}\\[1pt]
{\small\makebox[0.333\linewidth]{(a) Walk}\makebox[0.333\linewidth]{(b) Jump-A}\makebox[0.333\linewidth]{(c) Jump-B}}\\[3pt]
\includegraphics[width=\linewidth]{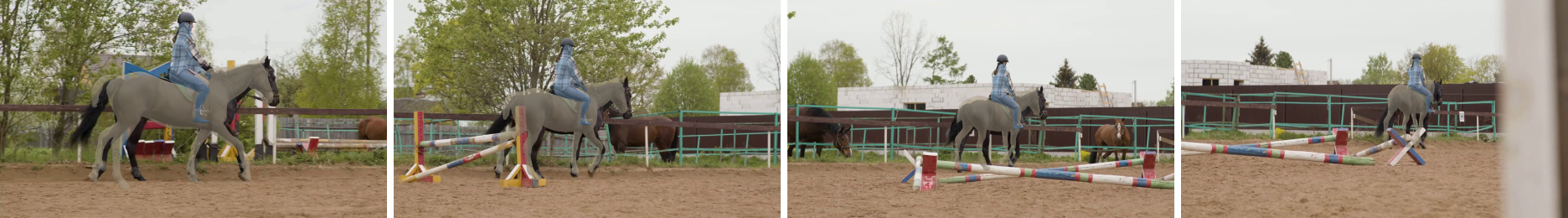}\\[1pt]
{\small (d) Walk, rendered through the estimated camera (four frames)}
\caption{\textbf{World frame.} (a)--(c): Walk, Jump-A and Jump-B in the levelled world frame, with the estimated camera (red) and the tracks of the horse (brown) and the rider (blue); earlier frames are translucent. (d): the world-frame reconstruction of Walk rendered through the estimated perspective camera on top of the input frames.}
\label{fig:world}
\end{figure*}

\paragraph{World frame.}
\Cref{tab:world,fig:world} summarize the world lifting. MoGe-2 estimates vertical fields of view of $21^\circ$ to $36^\circ$, and \cref{fig:fov} shows that this estimate matters: pan and travel change by a factor of two to four over plausible fields of view. On Walk the camera pans by $65^\circ$ while the horse walks 15.4\,m at 0.94\,m/s; on the jump approaches the camera turns by $5^\circ$ to $11^\circ$ and the horse covers 7 to 10\,m. On all riding clips the ground was constrained to contain the main direction of the hoof track, which is within $1^\circ$ of horizontal except on Jump-B ($11.6^\circ$). Rendering the world-frame result through the estimated perspective camera nearly reproduces the camera-frame fit on Walk (IoU$_h$/IoU$_m$ 0.556/0.469 vs.\ 0.565/0.497) and Jump-B (0.515/0.430 vs.\ 0.506/0.474), but loses agreement on the two Jump-A clips (0.453/0.425 vs.\ 0.540/0.540 and 0.438/0.430 vs.\ 0.477/0.505), where the horse approaches the camera quickly and the 0.6\,s smoothing is too strong. Stance hooves move at a speed comparable to the horse's (1.2 to 2.7\,m/s): the reconstructed legs swing but do not push backwards against the ground.

\subsection{Failure analysis}
\label{sec:failures}

\paragraph{Outside the task: a led horse.}
Lead was added on purpose to test the method outside its task. Nobody is mounted, but RoA applies its saddle and straddle terms regardless, as it is designed for riding. It misplaces the person relative to the horse and lowers the person's IoU from 0.71 for the baseline to 0.41 (\cref{tab:main} and last row of \cref{fig:qualitative}); here the baseline, which treats the person and the horse independently, is the better reconstruction: it keeps the person beside the horse (seat gap 134\,cm) with almost no penetration. The degradation on this clip is therefore expected and does not concern the riding task. The interaction terms should be gated by a test for riding, for instance based on the initial distance between pelvis and back and on the overlap of the person's mask with the horse's back. We report the case because applying a riding method to a led horse is the most likely way in which it will be misused.

\paragraph{Remaining penetration.}
With the collision term, 7.9\% of the rider's surface remains inside the horse on average, with P90 depths between 7.0\,cm (Walk) and 16.2\,cm (Jump-A long). Part of it is contact between the inner thighs and the barrel, which is partly physical since the saddle and soft tissue are not modelled, but depths beyond a few centimetres are errors: the bounded leg deformation cannot represent a bent knee wrapped around the barrel (\cref{sec:limitations}).

\section{Limitations and Future Work}
\label{sec:limitations}

\paragraph{Evaluation.}
No 3D ground truth is available for our clips. Our metrics measure plausibility (penetration, drift and world-frame consistency) and agreement with automatically computed masks, not accuracy, and five clips from four videos do not establish generalization. Calibrated data would allow accuracy to be measured directly. The PFERD dataset~\cite{pferd} provides synchronized video from ten cameras, 3D motion of more than 100 skin-attached markers and fits of the hSMAL model, and could be used to validate the horse side of the reconstruction. It is focused on the horse, however, and contains few ridden sequences with annotated riders, so validating the rider and the contact would require additional recordings in which both subjects are annotated.

\paragraph{Output parameterization.}
The output of RoA is a bounded deformation of the estimators' meshes, not a mesh re-decoded from the MHR and VAREN parameter spaces; a re-decoding stage that we tested reintroduced collisions and loosened the saddle contact. Applying the interaction terms directly to the body-model parameters, including the knee angles, is left to future work.

\paragraph{Depth along the camera axis.}
Weak perspective does not observe depth, and the interaction terms constrain it only along the horse's lateral direction. When the horse faces the camera, its longitudinal axis coincides with the camera axis and only the seat terms constrain the rider's fore--aft position. On the jump approaches, RoA reduces the forward drift, but it remains at 4.6 to 6.0\,cm, compared with 1.2\,cm on the walk seen from the side (\cref{tab:main}).

\paragraph{World lifting.}
The world frame assumes a camera that only rotates, a fixed focal length and a field of view estimated from single images. Camera translation, zoom and depth errors appear as ground tilt, speed errors and stance slip, which we report but do not correct. Stance slip is mainly due to the estimated limb motion, and a constraint that keeps stance hooves static would be a natural extension. The body orientation of the horse predicted by the estimator also jitters by $0.5^\circ$ to $8^\circ$ per frame (\cref{tab:world}); RoA does not modify it. The 0.6\,s smoothing of the world position is too strong for a horse that approaches the camera quickly, and an adaptive window would reduce the reprojection loss on the jump clips.

\paragraph{Other mounts.}
The methodology applies to any ridden animal for which a mesh model with fixed topology and a pose estimator exist; only the canonical regions and the body frame have to be defined. Such models are currently available for few ridden species besides horses (generic quadruped models such as SMAL~\cite{smal} are a possible starting point), and we have evaluated the method on horses only.

\paragraph{Broader impact.}
The reconstructions could support coaching and research on equine locomotion and welfare, but should not be used for veterinary, safety or judging decisions without validated data and expert involvement, and videos of identifiable riders require appropriate consent and usage rights.

\section{Conclusion}
\label{sec:conclusion}

We presented RoA, a general, training-free methodology for reconstructing a rider and a ridden animal from monocular video, instantiated for horseback riding. Compared with the raw outputs of the two estimators, RoA places the rider on the horse, removes most of the seat drift, reduces the penetration depth, makes the rider's motion relative to the horse considerably smoother, stays aligned with the input video, and places both subjects in a world frame. The reconstructions are visually far more plausible than the raw outputs; measuring their 3D accuracy requires calibrated recordings of ridden animals and is the main open task.

{
    \small
    \bibliographystyle{ieeenat_fullname}
    \bibliography{main}

@String(CVPR= {IEEE Conf. Comput. Vis. Pattern Recog.})

@String(ICCV= {Int. Conf. Comput. Vis.})

@String(ECCV= {Eur. Conf. Comput. Vis.})

@String(ACCV  = {ACCV})

@String(ICLR = {Int. Conf. Learn. Represent.})

@String(IJCAI = {IJCAI})

@article{sam3dbody,
  title={{SAM} {3D} Body: Robust Full-Body Human Mesh Recovery},
  author={Yang, Xitong and Kukreja, Devansh and Pinkus, Don and Sagar, Anushka and Fan, Taosha and Park, Jinhyung and Shin, Soyong and Cao, Jinkun and Liu, Jiawei and Ugrinovic, Nicolas and Feiszli, Matt and Malik, Jitendra and Doll{\'a}r, Piotr and Kitani, Kris},
  journal={arXiv preprint arXiv:2602.15989},
  year={2026}
}

@article{4dequine,
  title={{4DEquine}: Disentangling Motion and Appearance for {4D} Equine Reconstruction from Monocular Video},
  author={Lyu, Jin and An, Liang and Cheng, Pujin and Liu, Yebin and Tang, Xiaoying},
  journal={arXiv preprint arXiv:2603.10125},
  year={2026}
}

@inproceedings{dessie,
  title={Dessie: Disentanglement for Articulated {3D} Horse Shape and Pose Estimation from Images},
  author={Li, Ci and Yang, Yi and Weng, Zehang and Hernlund, Elin and Zuffi, Silvia and Kjellstr{\"o}m, Hedvig},
  booktitle={ACCV},
  year={2024}
}

@inproceedings{varen,
  title={{VAREN}: Very Accurate and Realistic Equine Network},
  author={Zuffi, Silvia and Mellbin, Ylva and Li, Ci and Hoeschle, Markus and Kjellstr{\"o}m, Hedvig and Polikovsky, Senya and Hernlund, Elin and Black, Michael J.},
  booktitle={CVPR},
  year={2024},
  pages={5374--5383}
}

@article{smpl,
  title={{SMPL}: A Skinned Multi-Person Linear Model},
  author={Loper, Matthew and Mahmood, Naureen and Romero, Javier and Pons-Moll, Gerard and Black, Michael J.},
  journal={ACM Trans. Graph.},
  volume={34},
  number={6},
  pages={248:1--248:16},
  year={2015}
}

@inproceedings{smplx,
  title={Expressive Body Capture: {3D} Hands, Face, and Body from a Single Image},
  author={Pavlakos, Georgios and Choutas, Vasileios and Ghorbani, Nima and Bolkart, Timo and Osman, Ahmed A. A. and Tzionas, Dimitrios and Black, Michael J.},
  booktitle={CVPR},
  pages={10967--10977},
  year={2019}
}

@inproceedings{smplify,
  title={Keep it {SMPL}: Automatic Estimation of {3D} Human Pose and Shape from a Single Image},
  author={Bogo, Federica and Kanazawa, Angjoo and Lassner, Christoph and Gehler, Peter and Romero, Javier and Black, Michael J.},
  booktitle={ECCV},
  year={2016}
}

@inproceedings{hmr,
  title={End-to-end Recovery of Human Shape and Pose},
  author={Kanazawa, Angjoo and Black, Michael J. and Jacobs, David W. and Malik, Jitendra},
  booktitle={CVPR},
  year={2018},
  pages={7122--7131}
}

@inproceedings{smal,
  title={{3D} Menagerie: Modeling the {3D} Shape and Pose of Animals},
  author={Zuffi, Silvia and Kanazawa, Angjoo and Jacobs, David W. and Black, Michael J.},
  booktitle={CVPR},
  pages={5524--5532},
  year={2017}
}

@article{hsmal,
  title={{hSMAL}: Detailed Horse Shape and Pose Reconstruction for Motion Pattern Recognition},
  author={Li, Ci and Ghorbani, Nima and Broom{\'e}, Sofia and Rashid, Maheen and Black, Michael J. and Hernlund, Elin and Kjellstr{\"o}m, Hedvig and Zuffi, Silvia},
  journal={arXiv preprint arXiv:2106.10102},
  year={2021},
  note={CV4Animals Workshop at CVPR 2021}
}

@inproceedings{barc,
  title={{BARC}: Learning to Regress {3D} Dog Shape from Images by Exploiting Breed Information},
  author={Rueegg, Nadine and Zuffi, Silvia and Schindler, Konrad and Black, Michael J.},
  booktitle={CVPR},
  year={2022}
}

@inproceedings{phosa,
  title={Perceiving {3D} Human-Object Spatial Arrangements from a Single Image in the Wild},
  author={Zhang, Jason Y. and Pepose, Sam and Joo, Hanbyul and Ramanan, Deva and Malik, Jitendra and Kanazawa, Angjoo},
  booktitle={ECCV},
  year={2020}
}

@inproceedings{prox,
  title={Resolving {3D} Human Pose Ambiguities with {3D} Scene Constraints},
  author={Hassan, Mohamed and Choutas, Vasileios and Tzionas, Dimitrios and Black, Michael J.},
  booktitle={ICCV},
  year={2019}
}

@inproceedings{jiang2020coherent,
  title={Coherent Reconstruction of Multiple Humans from a Single Image},
  author={Jiang, Wen and Kolotouros, Nikos and Pavlakos, Georgios and Zhou, Xiaowei and Daniilidis, Kostas},
  booktitle={CVPR},
  year={2020}
}

@inproceedings{behave,
  title={{BEHAVE}: Dataset and Method for Tracking Human Object Interactions},
  author={Bhatnagar, Bharat Lal and Xie, Xianghui and Petrov, Ilya A. and Sminchisescu, Cristian and Theobalt, Christian and Pons-Moll, Gerard},
  booktitle={CVPR},
  year={2022}
}

@inproceedings{hi4d,
  title={{Hi4D}: {4D} Instance Segmentation of Close Human Interaction},
  author={Yin, Yifei and Guo, Chen and Kaufmann, Manuel and Zarate, Juan Jose and Song, Jie and Hilliges, Otmar},
  booktitle={CVPR},
  year={2023},
  pages={17016--17027}
}

@inproceedings{glamr,
  title={{GLAMR}: Global Occlusion-Aware Human Mesh Recovery with Dynamic Cameras},
  author={Yuan, Ye and Iqbal, Umar and Molchanov, Pavlo and Kitani, Kris and Kautz, Jan},
  booktitle={CVPR},
  year={2022}
}

@inproceedings{slahmr,
  title={Decoupling Human and Camera Motion from Videos in the Wild},
  author={Ye, Vickie and Pavlakos, Georgios and Malik, Jitendra and Kanazawa, Angjoo},
  booktitle={CVPR},
  year={2023}
}

@inproceedings{wham,
  title={{WHAM}: Reconstructing World-grounded Humans with Accurate {3D} Motion},
  author={Shin, Soyong and Kim, Juyong and Halilaj, Eni and Black, Michael J.},
  booktitle={CVPR},
  year={2024},
  pages={2070--2080}
}

@inproceedings{tram,
  title={{TRAM}: Global Trajectory and Motion of {3D} Humans from in-the-wild Videos},
  author={Wang, Yufu and Wang, Ziyun and Liu, Lingjie and Daniilidis, Kostas},
  booktitle={ECCV},
  year={2024},
  pages={467--487}
}

@inproceedings{moge,
  title={{MoGe}: Unlocking Accurate Monocular Geometry Estimation for Open-Domain Images with Optimal Training Supervision},
  author={Wang, Ruicheng and Xu, Sicheng and Dai, Cassie and Xiang, Jianfeng and Deng, Yu and Tong, Xin and Yang, Jiaolong},
  booktitle={CVPR},
  year={2025},
  pages={5261--5271}
}

@inproceedings{moge2,
  title={{MoGe-2}: Accurate Monocular Geometry with Metric Scale and Sharp Details},
  author={Wang, Ruicheng and Xu, Sicheng and Dong, Yue and Deng, Yu and Xiang, Jianfeng and Lv, Zelong and Sun, Guangzhong and Tong, Xin and Yang, Jiaolong},
  booktitle={NeurIPS},
  volume={38},
  pages={40377--40408},
  year={2025}
}

@article{jacobson2013winding,
  title={Robust Inside-Outside Segmentation using Generalized Winding Numbers},
  author={Jacobson, Alec and Kavan, Ladislav and Sorkine-Hornung, Olga},
  journal={ACM Trans. Graph.},
  volume={32},
  number={4},
  year={2013}
}

@article{vanoosterom1983solid,
  title={The Solid Angle of a Plane Triangle},
  author={Van Oosterom, A. and Strackee, J.},
  journal={IEEE Trans. Biomed. Eng.},
  volume={BME-30},
  number={2},
  pages={125--126},
  year={1983}
}

@inproceedings{adam,
  title={Adam: A Method for Stochastic Optimization},
  author={Kingma, Diederik P. and Ba, Jimmy},
  booktitle={ICLR},
  year={2015}
}

@inproceedings{lucaskanade,
  title={An Iterative Image Registration Technique with an Application to Stereo Vision},
  author={Lucas, Bruce D. and Kanade, Takeo},
  booktitle={IJCAI},
  year={1981}
}

@article{ransac,
  title={Random Sample Consensus: A Paradigm for Model Fitting with Applications to Image Analysis and Automated Cartography},
  author={Fischler, Martin A. and Bolles, Robert C.},
  journal={Commun. ACM},
  volume={24},
  number={6},
  pages={381--395},
  year={1981}
}

@book{hartley2004multiple,
  title={Multiple View Geometry in Computer Vision},
  author={Hartley, Richard and Zisserman, Andrew},
  edition={2nd},
  publisher={Cambridge University Press},
  year={2004}
}

@article{savitzky1964smoothing,
  title={Smoothing and Differentiation of Data by Simplified Least Squares Procedures},
  author={Savitzky, Abraham and Golay, Marcel J. E.},
  journal={Analytical Chemistry},
  volume={36},
  number={8},
  pages={1627--1639},
  year={1964}
}

@misc{yolo11,
  title={Ultralytics {YOLO11}},
  author={Jocher, Glenn and Qiu, Jing},
  note={Version 11.0.0},
  howpublished={\url{https://github.com/ultralytics/ultralytics}},
  year={2024}
}

@article{pferd,
  title={The Poses for Equine Research Dataset ({PFERD})},
  author={Li, Ci and Mellbin, Ylva and Krogager, Johanna and Polikovsky, Senya and Holmberg, Martin and Ghorbani, Nima and Black, Michael J. and Kjellstr{\"o}m, Hedvig and Zuffi, Silvia and Hernlund, Elin},
  journal={Scientific Data},
  volume={11},
  pages={497},
  year={2024},
  doi={10.1038/s41597-024-03312-1}
}
}

\clearpage
\maketitlesupplementary
\appendix

\section{Implementation details}
\label{sec:impl}

\paragraph{Weights and optimization.}
\Cref{tab:weights} lists every loss weight. The same values are used for all clips and variants; an ablation sets the listed weight(s) to zero and changes nothing else. We run Adam (learning rate 0.015) for 300 iterations over all frames of a clip jointly. The winding-number inside test is refreshed every 10 iterations; the collision term uses 900 rider and 4000 horse surface samples, the image terms 900 area-weighted samples per subject, and the z-buffer runs at $48\times85$ in frame chunks with activation checkpointing. Runtimes are measured on one RTX~3090.

\begin{table}[h]
\centering\scriptsize\setlength{\tabcolsep}{3pt}
\caption{Loss weights (\texttt{configs/default.yaml}; $\lambda_{\rm occ}$ is a default of the code).}
\label{tab:weights}
\begin{tabular}{lr@{\hspace{12pt}}lr}
\toprule
Term & Weight & Term & Weight\\
\midrule
image $\lambda_{\rm img}$ & 12 & collision $\lambda_{\rm col}$ & 400\\
occlusion $\lambda_{\rm occ}$ & 20 & straddle $\lambda_{\rm str}$ & 120\\
contact distance & 16 & centre line & 180\\
saddle offset & 180 & balance & 30\\
normal/tangent & 120 & body ellipsoid & 32\\
centre coupling & 120 & temporal & 40\\
relative velocity & 18 & anchor & 0.25\\
\bottomrule
\end{tabular}
\end{table}

\paragraph{World lifting.}
Corners from \texttt{goodFeaturesToTrack} (1200, quality 0.01, distance 7\,px) outside 31\,px-dilated instance masks; pyramidal Lucas--Kanade~\cite{lucaskanade} (window 21, 3 levels); homography by RANSAC~\cite{ransac} with a 2\,px threshold, accepted with at least 25 inliers (otherwise the rotation increment is the identity). Savitzky--Golay smoothing~\cite{savitzky1964smoothing} of order 2 over 0.6\,s. MoGe-2 (\texttt{moge-2-vitl-normal}) runs on 8 frames.

\paragraph{Coordinate conventions.}
VAREN predictions are in OpenCV camera axes ($+Y$ down, $+Z$ away from the camera) and are rotated by $180^\circ$ about $X$; SAM 3D Body outputs are already $+Y$ up. In the resulting frame $+Z$ points towards the camera. We verified this by checking that both bodies have the same handedness (rider and horse face the same direction) and that the horse's predicted translation lies in front of the camera.

\section{Per-clip ablations}
\label{sec:perclip}
\Cref{tab:perclip} breaks \cref{tab:ablation} down by clip, including Lead, which is excluded from the averages.

\begin{table}[t]
\centering\scriptsize\setlength{\tabcolsep}{2.0pt}
\caption{\textbf{Per-clip results} (columns as in \cref{tab:ablation}). Bold: best value per clip and column, ties included.}
\label{tab:perclip}
\begin{tabular}{lccccccc}
\toprule
Variant & IoU$_h$ & IoU$_m$ & Inside & P90 & Gap & Fwd. & Lat.\\
 & $\uparrow$ & $\uparrow$ & \%\,$\downarrow$ & cm\,$\downarrow$ & cm\,$\downarrow$ & cm\,$\downarrow$ & cm\,$\downarrow$\\
\midrule
\multicolumn{8}{l}{\textit{Walk}}\\
\quad Baseline & \textbf{0.667} & \textbf{0.517} & 12.0 & 28.2 & 10.4 & 8.7 & 17.1\\
\quad w/o joint opt. & 0.628 & 0.515 & 15.6 & 21.6 & 1.7 & 4.6 & \textbf{1.2}\\
\quad w/o contact & 0.558 & 0.495 & 10.0 & 7.1 & 1.1 & 1.6 & 2.2\\
\quad w/o image & 0.537 & 0.500 & 7.9 & 6.3 & 0.9 & 1.3 & 2.6\\
\quad world axes & 0.566 & \textbf{0.517} & \textbf{3.6} & \textbf{5.0} & 1.5 & 3.7 & 1.5\\
\quad w/o collision & 0.567 & 0.513 & 19.0 & 18.9 & 0.9 & 1.3 & 2.3\\
\quad Full RoA & 0.565 & 0.497 & 9.4 & 7.0 & \textbf{0.8} & \textbf{1.2} & 2.5\\
\midrule
\multicolumn{8}{l}{\textit{Jump-A}}\\
\quad Baseline & \textbf{0.565} & 0.499 & 9.7 & 24.3 & 16.0 & 13.2 & 5.9\\
\quad w/o joint opt. & 0.526 & 0.507 & 8.7 & 9.4 & 4.0 & 9.2 & 3.0\\
\quad w/o contact & 0.547 & 0.537 & 8.0 & 11.1 & 1.4 & 4.6 & 1.4\\
\quad w/o image & 0.439 & 0.483 & 6.8 & 10.3 & 1.1 & 5.1 & \textbf{1.1}\\
\quad world axes & 0.522 & \textbf{0.548} & \textbf{4.4} & \textbf{7.6} & 1.8 & \textbf{3.6} & 1.7\\
\quad w/o collision & 0.545 & 0.534 & 8.2 & 11.5 & 1.1 & 4.6 & 1.4\\
\quad Full RoA & 0.540 & 0.540 & 6.7 & 9.9 & \textbf{1.0} & 4.9 & 1.3\\
\midrule
\multicolumn{8}{l}{\textit{Jump-A long}}\\
\quad Baseline & \textbf{0.539} & 0.503 & 12.0 & 26.9 & 13.5 & 10.1 & 10.5\\
\quad w/o joint opt. & 0.461 & 0.504 & 16.9 & 24.3 & 4.6 & 11.6 & 3.8\\
\quad w/o contact & 0.480 & 0.504 & 8.6 & 16.2 & 1.2 & 5.1 & 2.4\\
\quad w/o image & 0.398 & 0.500 & 8.3 & 16.3 & \textbf{1.0} & 4.6 & \textbf{2.2}\\
\quad world axes & 0.479 & \textbf{0.507} & \textbf{5.7} & \textbf{7.2} & 3.1 & \textbf{4.0} & 2.4\\
\quad w/o collision & 0.477 & 0.502 & 9.4 & 16.2 & \textbf{1.0} & 4.7 & 2.3\\
\quad Full RoA & 0.477 & 0.505 & 8.8 & 16.2 & \textbf{1.0} & 4.6 & 2.4\\
\midrule
\multicolumn{8}{l}{\textit{Jump-B}}\\
\quad Baseline & \textbf{0.606} & 0.447 & 14.6 & 30.5 & 19.6 & 14.3 & 36.0\\
\quad w/o joint opt. & 0.568 & 0.449 & 17.1 & 26.0 & 7.2 & 6.1 & \textbf{0.7}\\
\quad w/o contact & 0.510 & 0.475 & 7.3 & 12.2 & \textbf{1.3} & 8.8 & 1.3\\
\quad w/o image & 0.437 & 0.453 & \textbf{6.4} & 15.1 & 1.5 & 5.6 & 1.7\\
\quad world axes & 0.509 & 0.473 & 6.5 & \textbf{12.1} & 1.5 & \textbf{4.6} & 3.8\\
\quad w/o collision & 0.520 & \textbf{0.476} & 8.7 & 14.8 & 1.6 & 5.7 & 1.6\\
\quad Full RoA & 0.506 & 0.474 & 6.7 & 13.7 & 1.4 & 6.0 & 1.6\\
\midrule
\multicolumn{8}{l}{\textit{Lead$^\dagger$}}\\
\quad Baseline & \textbf{0.709} & \textbf{0.469} & \textbf{0.0} & \textbf{0.0} & 134.3 & 9.8 & 51.1\\
\quad w/o joint opt. & 0.644 & 0.468 & 0.4 & 3.2 & 130.4 & 11.2 & 23.5\\
\quad w/o contact & 0.466 & 0.416 & 2.6 & 9.3 & 108.0 & 11.0 & 9.5\\
\quad w/o image & 0.167 & 0.445 & 3.3 & 10.4 & 97.7 & 10.5 & \textbf{7.3}\\
\quad world axes & 0.432 & 0.402 & 4.3 & 10.7 & \textbf{94.6} & \textbf{8.8} & 12.6\\
\quad w/o collision & 0.406 & 0.396 & 3.3 & 10.5 & 98.1 & 10.2 & 7.5\\
\quad Full RoA & 0.406 & 0.396 & 3.4 & 10.5 & 98.1 & 10.2 & 7.5\\
\bottomrule
\end{tabular}
\end{table}

\end{document}